\documentclass[sigconf]{acmart}

\usepackage{booktabs} % For formal tables
\usepackage{diagbox}
\usepackage{stackengine}
\usepackage{tikz}

\setcopyright{none}

\setcitestyle{square}
\AtBeginDocument{%
  \providecommand\BibTeX{{%
    \normalfont B\kern-0.5em{\scshape i\kern-0.25em b}\kern-0.8em\TeX}}}

\begin{document}

\author{Jingsen Zhu}
\email{zhujingsen@zju.edu.cn}
\affiliation{%
% \email{zhujingsen\@zju.edu.cn}
\institution{State Key Lab of CAD\&CG, Zhejiang University}
% \city{Hangzhou}
\country{China}}

\author{Fujun Luan}
\email{fluan@adobe.com}
\affiliation{%
\institution{Adobe Research}
% \city{San Jose}
\country{USA}}

\author{Yuchi Huo}
\authornote{Denotes corresponding author.}
\email{huo.yuchi.sc@gmail.com}
\affiliation{%
\institution{State Key Lab of CAD\&CG, Zhejiang University}
\institution{Zhejiang Lab}
% \city{Hangzhou}
\country{China}}

\author{Zihao Lin}
% \authornotemark[1]
\email{zihaolin@zju.edu.cn}
\affiliation{%
\institution{State Key Lab of CAD\&CG, Zhejiang University}
% \city{Hangzhou}
\country{China}}

\author{Zhihua Zhong}
% \authornotemark[1]
\email{zhongzhihua@zju.edu.cn}
\affiliation{%
\institution{State Key Lab of CAD\&CG, Zhejiang University}
% \city{Hangzhou}
\country{China}}

\author{Dianbing Xi}
% \authornotemark[1]
\email{db.xi@zju.edu.cn}
\affiliation{%
\institution{State Key Lab of CAD\&CG, Zhejiang University}
% \city{Hangzhou}
\country{China}}

\author{Jiaxiang Zheng}
\email{xuanfeng@qunhemail.com}
\affiliation{%
\institution{KooLab, Manycore}
% \city{Hangzhou}
\country{China}}

\author{Rui Tang}
\email{ati@qunhemail.com}
\affiliation{%
\institution{KooLab, Manycore}
% \city{Hangzhou}
\country{China}}

% \author{Wei Hua}
% \email{huawei@zhejianglab.com}
% \affiliation{%
% \institution{Zhejiang Lab}
% \country{China}}

\author{Hujun Bao}
% \authornotemark[1]
\email{bao@cad.zju.edu.cn}
\affiliation{%
\institution{State Key Lab of CAD\&CG, Zhejiang University}
% \city{Hangzhou}
\country{China}}

\author{Rui Wang}
\authornotemark[1]
\email{rwang@cad.zju.edu.cn}
\affiliation{%
\institution{State Key Lab of CAD\&CG, Zhejiang University}
% \city{Hangzhou}
\country{China}}

%%
%% The "title" command has an optional parameter,
%% allowing the author to define a "short title" to be used in page headers.
\title[Learning-based Inverse Rendering of Complex Indoor Scenes with Differentiable MC Raytracing]{Learning-based Inverse Rendering of Complex Indoor Scenes with Differentiable Monte Carlo Raytracing}

%%
%% The "author" command and its associated commands are used to define
%% the authors and their affiliations.
%% Of note is the shared affiliation of the first two authors, and the
%% "authornote" and "authornotemark" commands
%% used to denote shared contribution to the research.

%%
%% By default, the full list of authors will be used in the page
%% headers. Often, this list is too long, and will overlap
%% other information printed in the page headers. This command allows
%% the author to define a more concise list
%% of authors' names for this purpose.
\renewcommand{\shortauthors}{Zhu, et~al.}

%%
%% The abstract is a short summary of the work to be presented in the
%% article.
\begin{abstract}
Indoor scenes typically exhibit complex, spatially-varying appearance from global illumination, making inverse rendering a challenging ill-posed problem. This work presents an end-to-end, learning-based inverse rendering framework incorporating differentiable Monte Carlo raytracing with importance sampling. The framework takes a single image as input to jointly recover the underlying geometry, spatially-varying lighting, and photorealistic materials. Specifically, we introduce a physically-based differentiable rendering layer with screen-space ray tracing, resulting in more realistic specular reflections that match the input photo. In addition, we create a large-scale, photorealistic indoor scene dataset with significantly richer details like complex furniture and dedicated decorations. Further, we design a novel out-of-view lighting network with uncertainty-aware refinement leveraging hypernetwork-based neural radiance fields to predict lighting outside the view of the input photo. Through extensive evaluations on common benchmark datasets, we demonstrate superior inverse rendering quality of our method compared to state-of-the-art baselines, enabling various applications such as complex object insertion and material editing with high fidelity. Code and data will be made available at \url{https://jingsenzhu.github.io/invrend}.
\end{abstract}



%keywords

%% A "teaser" image appears between the author and affiliation
%% information and the body of the document, and typically spans the
%% page.
\begin{teaserfigure}
  \setlength\tabcolsep{1.25pt}
  \vspace*{-1ex}
  \centering
  \begin{tabular}{cccc}
        \textsc{Holistic inverse rendering} & \textsc{Multiple complex object insertion}  &  \multicolumn{2}{c}{\textsc{ Material editing }} \\
       \includegraphics[height=0.19\textwidth]{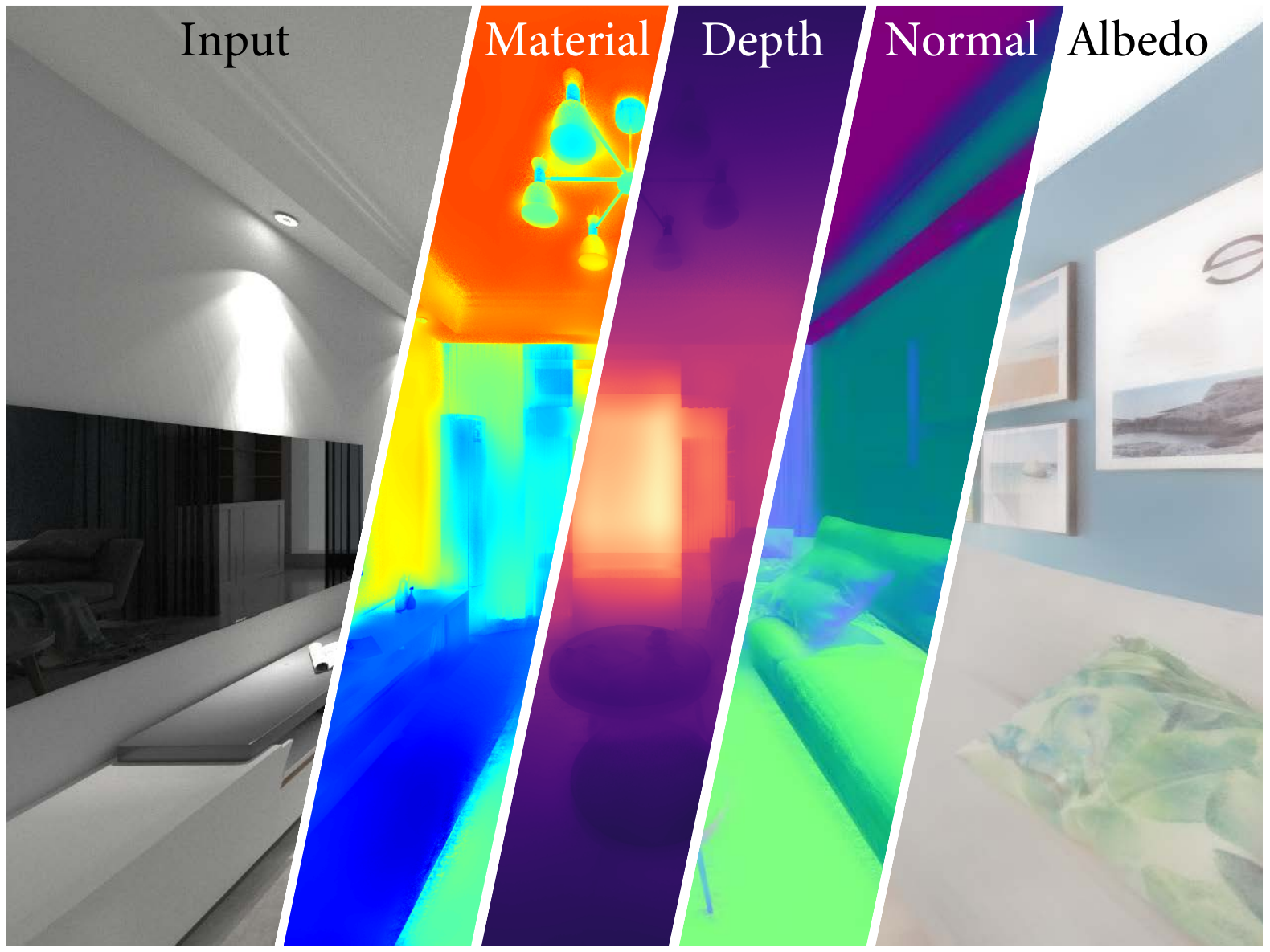}& \includegraphics[height=0.19\textwidth]{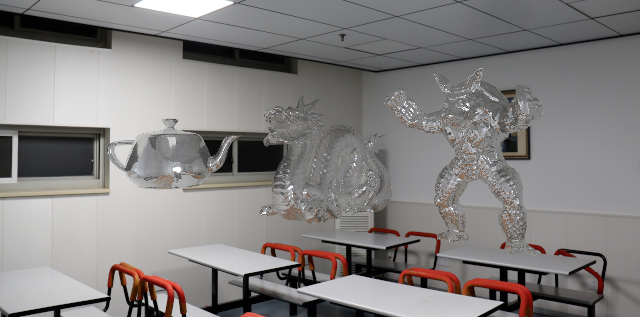} &\includegraphics[height=0.19\textwidth]{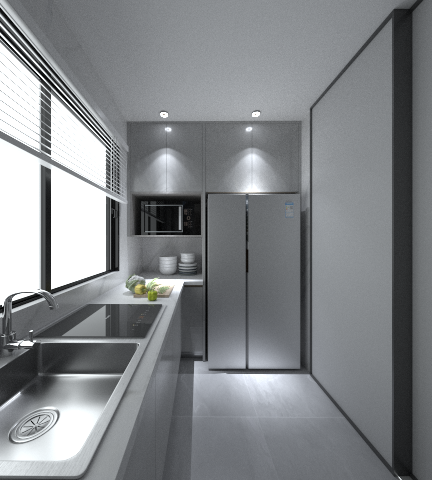}
       &\includegraphics[height=0.19\textwidth]{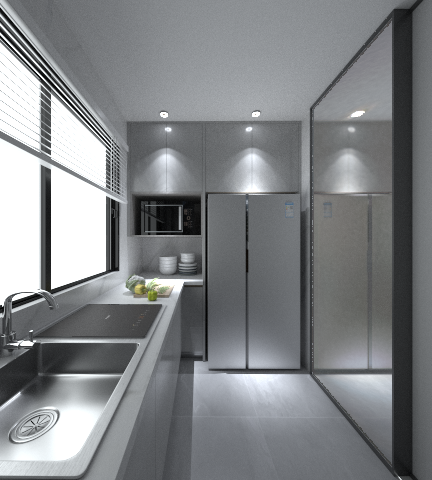}
  \end{tabular}
  \vspace*{-3ex}
  \caption{We present a learning-based approach for inverse rendering of complex indoor scenes with differentiable Monte Carlo raytracing. Our method takes a single indoor scene RGB image as input and automatically infers its underlying surface reflectance (represented by microfacet GGX), geometry, and spatially-varying illumination (first column). Consequently, this enables us to perform photorealistic editing of the scene, such as inserting multiple complex virtual objects (second column, note that the inserted models are highly glossy) and editing surface materials faithfully with global illumination (last two columns, note that the wall is modified to a mirror that correctly presents specular reflections of the kitchen, and the glossy cooktop is modified to Lambertian appearance).  }
  \label{fig:teaser}
%   \vspace*{-1ex}
\end{teaserfigure}

%%
%% This command processes the author and affiliation and title
%% information and builds the first part of the formatted document.
\maketitle

\section{Introduction}
Inverse rendering of complex indoor scenes has been a long-standing challenge in computer graphics and vision. Given a single real-world image, global illumination effects such as shadows, specular highlights, and glossy interreflections are baked into the observed pixel values, imposing a particularly difficult task of simultaneously recovering the underlying scene geometry, spatially-varying surface reflectance, and arbitrary unknown illumination. Traditional optimization-based approaches rely on dedicatedly designed regularization and hand-crafted priors to tackle this problem. Unfortunately, such methods often fail in real-world scenarios due to overly simplified assumptions, leading to noticeable artifacts in both the decomposition and re-rendered results.

On the other hand, recent advances in inverse rendering~\cite{wang2021learning,li2020inverse,srinivasan2020lighthouse} leveraging deep learning methods have demonstrated impressive results on such scene inference tasks, where the underlying physical priors are supposed to be learnt automatically through an offline, supervised training process, typically on a large-scale, synthetic or labeled real training dataset of complex indoor scenes. Note that, it is extremely difficult, if not impossible, to generate ground truth labels of spatially-varying illumination and materials of an arbitrary real-world scene, and hence one crucial keypoint to the success of such methods is the fidelity and photorealism of the synthetic training data. Another essential factor strongly influencing the inference accuracy is the network structure design. Intuitively speaking, since inverse rendering is inverting the physical light transport, a physically-based differentiable rendering layer regularizing the parameter space can act as a  meaningful prior, improving the robustness and generalization capability of the neural network regarding material and lighting decomposition, and thus also the geometry estimation in return. Consequently, the performance of these learning-based inverse rendering methods heavily depends on: 1) the quality of the training datasets, and 2) the design of the neural network architecture. 

To address the aforementioned challenges, we propose a novel Monte Carlo differentiable rendering layer with importance sampling to faithfully simulate the physical light transport of an indoor scene. Experiments show that this is especially helpful in restoring the specular reflections of a given scene, and our method produces much more realistic re-rendered results comparing previous baselines. Unlike previous work that directly uses the local feature at a ray-surface intersection point, our approach importance samples the local incident radiance field of it via screen space ray tracing (SSRT) and uncertainty-aware, hypernetwork-based out-of-view lighting estimation. To facilitate training, we introduce a large-scale (${\sim} 4000$) complex indoor scene dataset, \textsc{InteriorVerse}. As far as we know, our dataset contains the highest quality with rich details compared to existing indoor scene datasets (e.g., OpenRooms~\cite{li2021openrooms} or SUNCG~\cite{song2017semantic}), including complex furniture and dedicated decorations procedurally designed by professional digital artists, rendered with physically-based GGX model~\cite{walter2007microfacet} using a modern GPU-based path tracing engine.

Concretely, our contributions include: 
\begin{itemize}
    \item A learning-based monocular inverse rendering framework of complex indoor scenes that recovers albedo, surface normal, depth, metallic, and roughness from a single indoor scene image.
    \item A novel Monte Carlo differentiable rendering layer with importance sampling, which correctly estimates the local incident radiance field using screen space ray tracing.
    \item An uncertainty-aware out-of-view light network leveraging hypernetwork-based neural radiance fields for robust out-of-view lighting estimation.
    \item A high-quality, large-scale complex indoor scene dataset, \textsc{InteriorVerse}, that contains rich details with high fidelity.
\end{itemize}
% To validate our proposed inverse rendering pipeline, we compare with existing state-of-the-art baselines on a set of common benchmark datasets of indoor scenes and demonstrate that our method consistently improves the result, qualitatively and quantitatively. Our implementation will be made publicly available.

\section{Related Work}

\paragraph{Inverse Rendering of Indoor Scenes} Inverse rendering attempts to reconstruct geometry and spatially-varying material and lighting information from monocular (which is our case) or multiple RGB images. Most previous methods only recognize one or part of the above attributes. Geometry reconstructions, including depth estimation and surface normal reconstruction, has been widely studied \cite{eigen2015predicting,liu2019planercnn}.
Most material reconstruction methods are only able to either estimate diffuse albedo~\cite{li2018cgintrinsics, barron2013intrinsic, karsch2014automatic} or classify material categories~\cite{bell2015material}.
For lighting estimation, recent deep learning methods have made progress in estimating global~\cite{gardner2017learning,gardner2019deep} and even spatially-varying~\cite{garon2019fast,song2019neural,li2020inverse} lighting conditions.
Recent works attempt to predict multiple intrinsics jointly by a holistic inverse rendering framework. Li et al.~\shortcite{li2020inverse} proposed a method to reconstruct disentangled geometry, spatially-varying reflectance and lighting from a single RGB indoor scene image.

\paragraph{Lighting Estimation and Relighting.}
Light estimation is one of the sub-tasks of inverse rendering. Most previous works ignore spatially-varying effects and predict a single environment map for the whole scene \cite{gardner2017learning,sengupta2019neural,munkberg2022extracting}. Indoor scenes suffer from spatial variations, thus recent work explores spatially-varying lighting estimation for indoor scenes. The representation of spatially-varying illumination includes environment maps, per-pixel spherical lobes~\cite{li2020inverse} (spherical Harmonics/Gaussians), or 3D voxel grids~\cite{wang2021learning}. Relighting is also a widely-studied relevant  task. \citet{griffiths2022outcast} leverages screen-space method to detect occlusion and cast shadows to relight an outdoor image. \citet{li2022physically} proposed a novel pipeline to modify the light conditions within an indoor scene.

\paragraph{Neural Scene Representations.} 
Neural representations are a rapidly growing area of research. Recent advances include  voxels~\cite{yu2021plenoxels,sun2021direct}, hashgrids~\cite{muller2022instant}, point clouds~\cite{aliev2020neural}, and neural implicit functions~\cite{mildenhall2020nerf,wang2021neus,yariv2021volume,yariv2020multiview}. 
Neural radiance fields (NeRFs)~\cite{mildenhall2020nerf} represents scenes as neural implicit functions, encoding a scene as a continuous volumetric radiance field of color and density. With volume rendering, a NeRF can synthesize novel view images with promising results. Our proposed method uses a NeRF as the representation of the out-of-view area of the scene (Sec.~\ref{sec:background}).

\paragraph{Differentiable Rendering.} A number of recent inverse rendering works utilize differentiable rendering to recover complex light transport effects. Some recent works have proposed general-purpose physically-based differentiable renderers~\cite{Li:2018:DMC,NimierDavidVicini2019Mitsuba2}. \citet{Zhang:2020:PSDR} and \citet{Zeltner2021MonteCarlo} discussed a rigurous theory of differentiable light transport and Monte-Carlo combinations. These physically-based methods achieve high-quality global illumination effects at the cost of substantial performance overhead. Some differentiable rendering techniques are customized for specific purpose such as texture~\cite{nimier2021material}, split-sum lighting and mesh extraction~\cite{munkberg2022extracting}. Our method designs a Monte-Carlo based in-network differentiable rendering layer to recover the appearance of indoor scenes (Sec.~\ref{sec:render}).

\paragraph{Indoor Scene Datasets.} 
Supervised learning requires a large database of indoor scene images and their corresponding ground truth geometry, material, and lighting for network training. Datasets include 3D shape models~\cite{chang2015shapenet}, real-world scans~\cite{chang2017matterport3d, dai2017scannet}, and scene datasets~\cite{song2017semantic,savva2017minos,li2018interiornet,li2021openrooms}, which can be classified as either real or synthetic data. Real datasets provide real-world images and geometry, while synthetic datasets provide arbitrary scene annotations for inverse rendering, some of which, such as materials and illumination, are difficult to acquire from real world. To the best of our knowledge, InteriorNet~\cite{li2018interiornet} and OpenRooms~\cite{li2021openrooms} are so far the highest-quality public indoor datasets with spatially-varying photorealistic material and illumination annotations. Unfortunately, InteriorNet provides only LDR results, while OpenRooms provides only lighting information on the scene surface (instead of at any 3D location), and lacks the complexity of material and furniture variations. We present a new indoor scene HDR dataset to tackle their shortcomings.

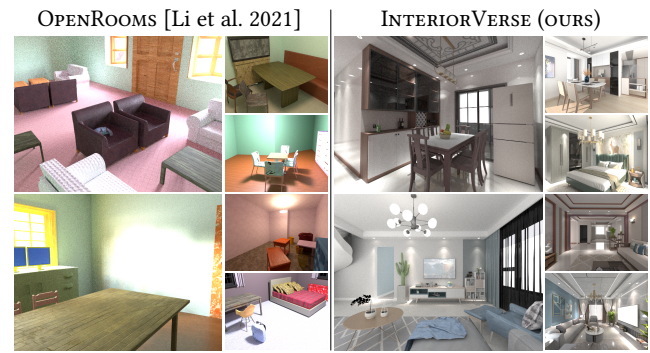
\begin{figure}[b]
    \centering
    \setlength{\fboxsep}{0.0pt}%
    \setlength{\fboxrule}{0.75pt}%
    \vspace*{-1.0\baselineskip}%
    \hspace*{-0.6ex}%
    \begin{tikzpicture}[x=0.075\textwidth, y=0.075\textwidth,every text node part/.style={align=center}]
    \node[anchor=north west] at (-0.03, 0) {\includegraphics[width=0.155\textwidth]{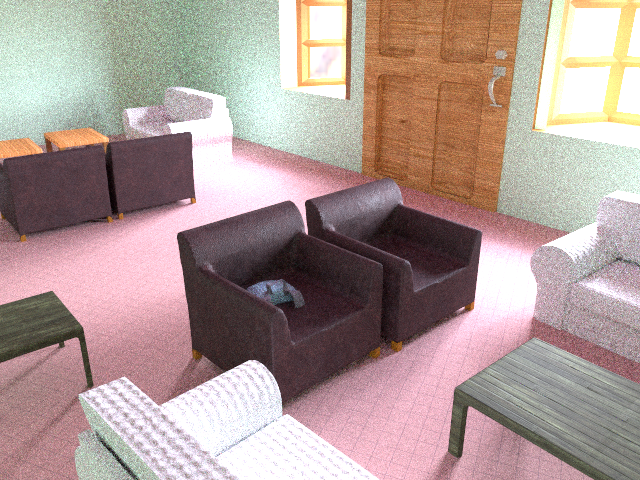}};
    \node[anchor=north west] at (3.16, 0) {\includegraphics[width=0.155\textwidth]{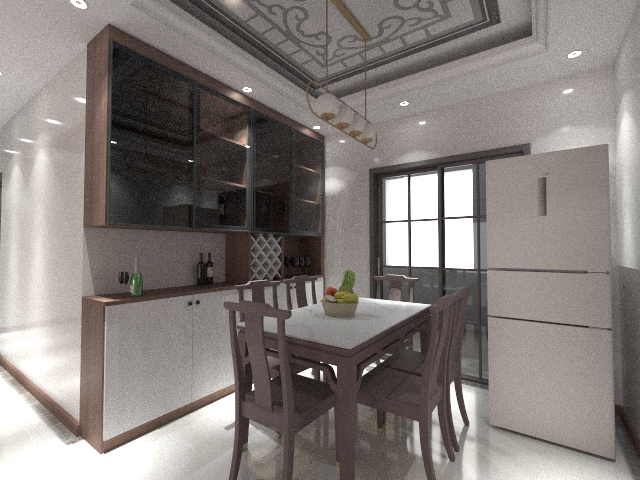}};
    
    \node[anchor=north west] at (2.07, 0) {\includegraphics[width=0.076\textwidth]{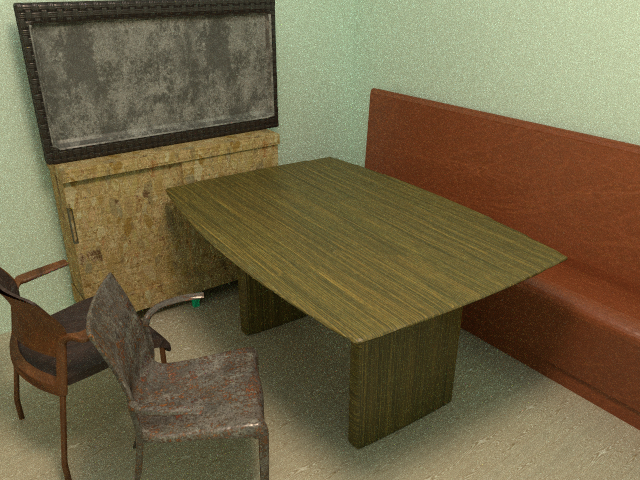}};
    \node[anchor=north west] at (2.07, -0.79+0.0021) {\includegraphics[width=0.076\textwidth]{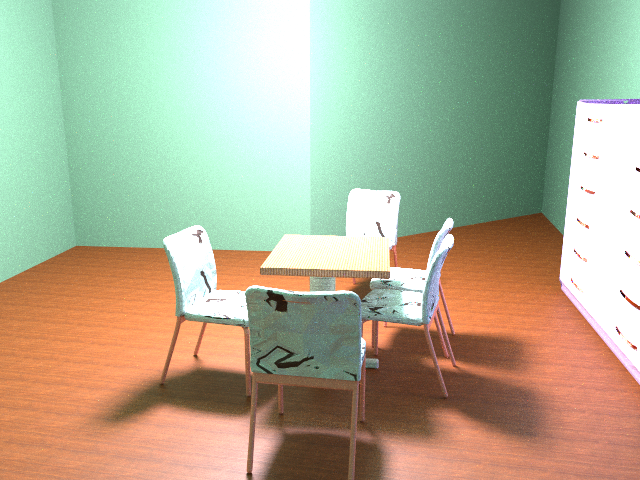}};
    \node[anchor=north west] at (2.07, -1.58) {\includegraphics[width=0.076\textwidth]{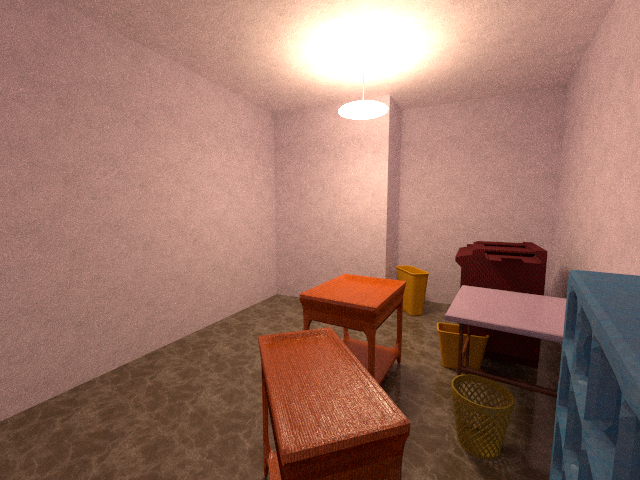}};
    \node[anchor=north west] at (2.07, -1.58-0.79+0.0021) {\includegraphics[width=0.076\textwidth]{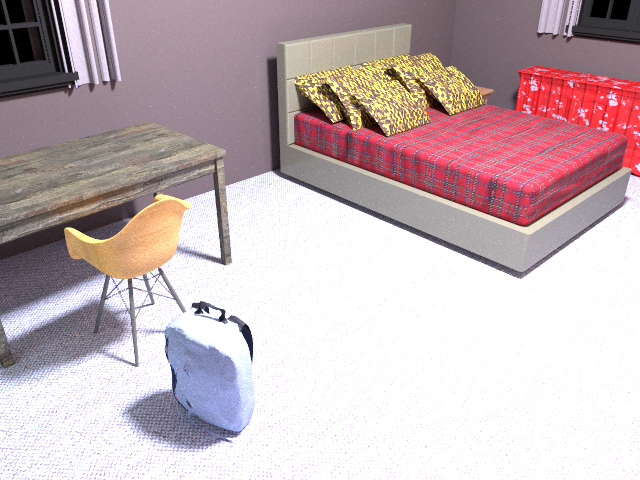}};
    
    \node[anchor=north west] at (-0.03, -1.58) {\includegraphics[width=0.155\textwidth]{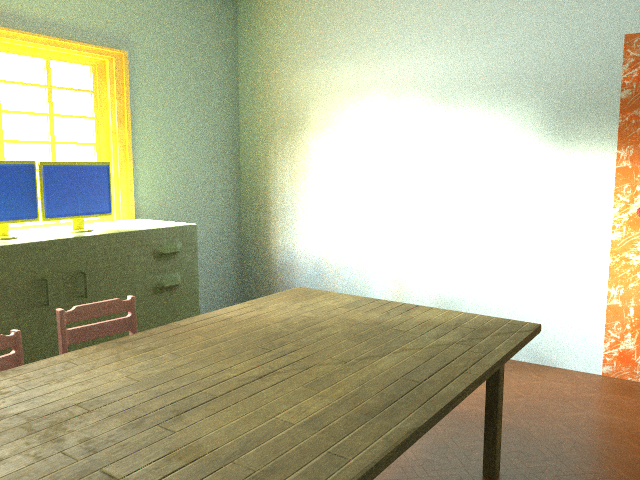}};
    \node[anchor=north west] at (3.16, -1.58) {\includegraphics[width=0.155\textwidth]{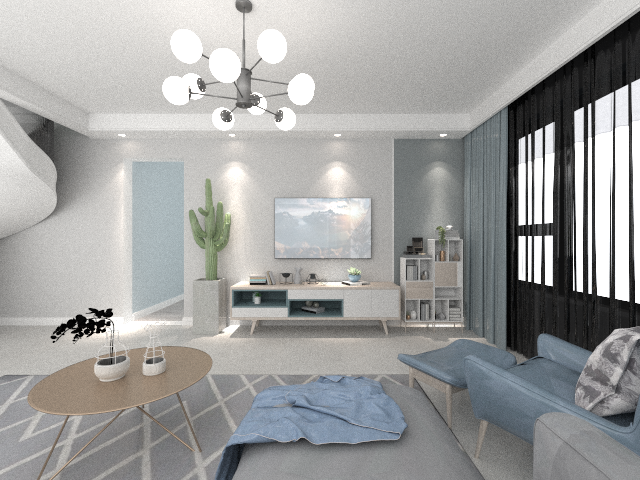}};
    
    \node[anchor=north west] at (5.26, 0) {\includegraphics[width=0.076\textwidth]{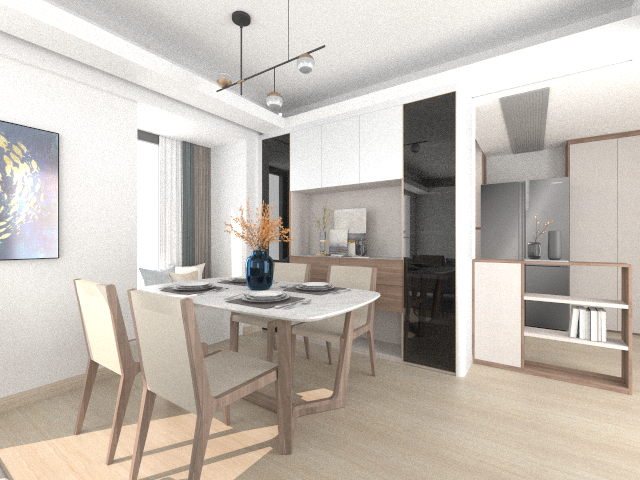}};
    \node[anchor=north west] at (5.26, -0.79+0.0021) {\includegraphics[width=0.076\textwidth]{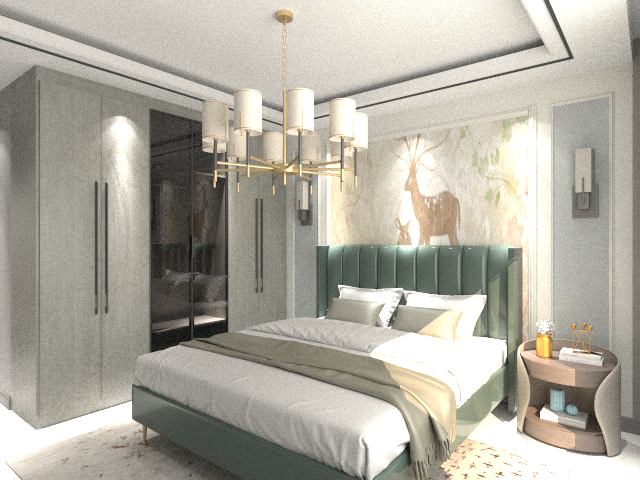}};
    \node[anchor=north west] at (5.26, -1.58) {\includegraphics[width=0.076\textwidth]{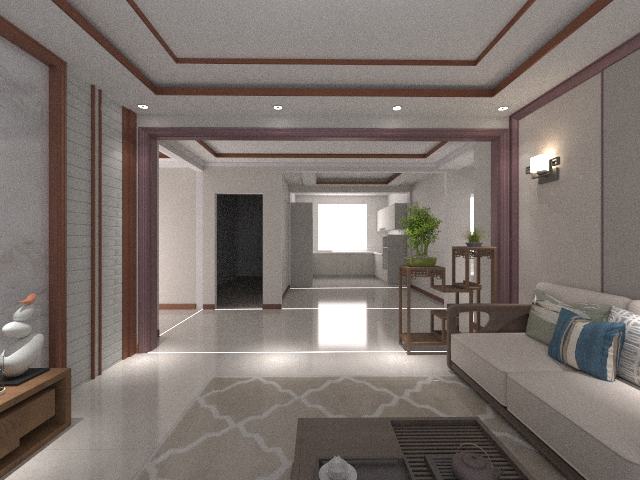}};
    \node[anchor=north west] at (5.26, -1.58-0.79+0.0021) {\includegraphics[width=0.076\textwidth]{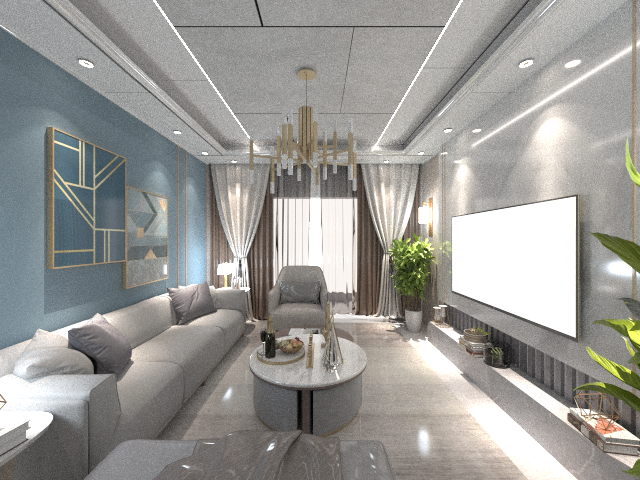}};
 
    \draw[ black, very thin] (3.205,0.18) -- (3.205,-3.218);
    \node at (1.6, 0.08){\textsc{OpenRooms}~\cite{li2021openrooms}};
    \node at (4.8, 0.08){\textsc{InteriorVerse (ours)}};

    \end{tikzpicture}%
    % \vspace*{-1.0\baselineskip}%
    \vspace*{-2.5ex}
    \caption{\textbf{Example dataset images} from OpenRooms~\cite{li2021openrooms} (left) and our \textsc{InteriorVerse} dataset (right). Note that our dataset contains more diversified geometry, material (especially glossy and specular BRDFs) and complex lighting conditions comparing OpenRooms. Zoom in for details.}
    \label{fig:dataset}
    % \vspace*{-1.\baselineskip}%
\end{figure}

\begin{figure*}[ht]
    \centering
    \includegraphics[width=\textwidth]{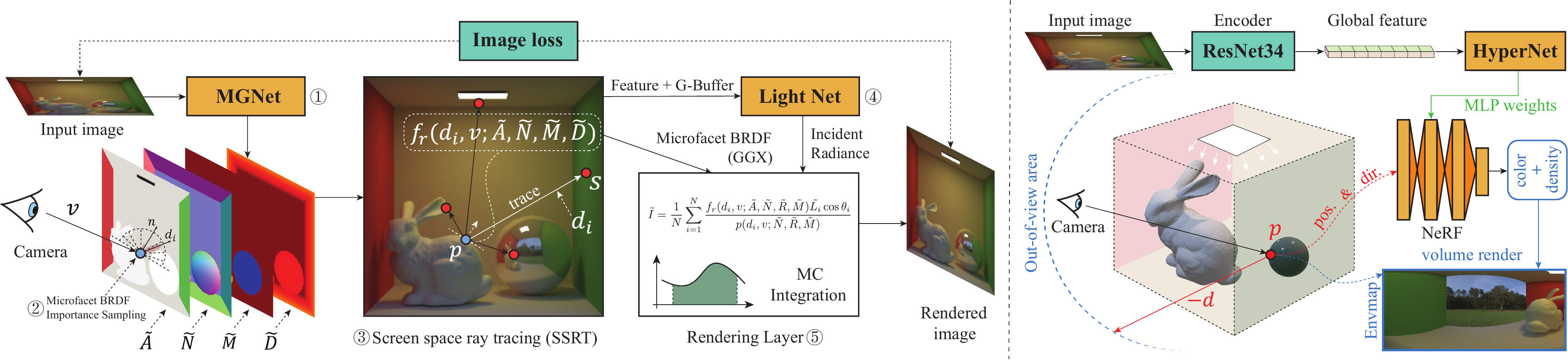}
    \vspace*{-3.5ex}
    \caption{\textbf{Overview of the pipeline}. On the left, we show the workflow throughout our inverse rendering framework: (i) The spatially-varying material and geometry maps are predicted by $\textbf{MGNet}$ (ii) According to the predicted material, geometry and view direction associated with each pixel point $\mathbf{p}$, a BRDF importance sampling is performed to generate per-pixel incident directions $\mathbf{d}_i$ (iii) We use screen-space raytracing to trace the source point $\mathbf{s}$ of the query light. The corresponding local feature vector is extracted from feature map $F$ via projection of point $\mathbf{s}$. (iv) The feature is passed to $\mathbf{LightNet}$ along with light direction and auxiliary G-Buffer information to predict the incident radiance $\tilde{L_i}$ (v) Monte-Carlo integration (Eq.~\ref{equ:render}) is used to calculate the final re-rendered result. On the right, we show our out-of-view light estimation. We use a hypernetwork to predict the weights of the NeRF MLP and volume render the background lighting.}
    \label{fig:pipeline}
    % \vspace*{-1.\baselineskip}%
    % \vspace{-1ex}
\end{figure*}

\section{InteriorVerse: A Large-Scale, Photorealistic Indoor Scene Dataset}\label{sec:data}

A high-quality dataset is crucial for learning-based inverse rendering. It's extremely difficult to acquire spatially-varying material and lighting ground truth in real world complex indoor scenes. Therefore, we render a synthetic dataset to supervise training. The SUNCG dataset~\cite{song2017semantic} is a manually-created large-scale dataset for indoor scenes, but they use non-physical Phong BRDF and render with OpenGL, which severely limits its photorealism. PBRS~\cite{zhang2017physically} and CG-PBR~\cite{sengupta2019neural} datasets are rendered with physically-based renderers, but both still use Phong BRDF and do not provide spatially-varying lighting ground truth for an arbitrary 3D location. InteriorNet~\cite{li2018interiornet} is a large-scale photorealistic indoor scene dataset providing multiple camera views and panoramas, but the images they provide are LDR, limiting the dynamic range of illumination. OpenRooms~\cite{li2021openrooms} is by far the only HDR dataset with spatially-varying lighting rendered using physically-based microfacet BRDF. However, as shown in Fig.~\ref{fig:dataset}, it presents overly simplified furniture models and layouts, insufficient material and lighting variations, leading to less faithful appearance comparing to real world data consequently.

In this work, we create a new high-quality indoor scene dataset called \textsc{InteriorVerse}, which has the following advantages in data quality over existing alternatives: (1) the scene layouts of our dataset have richer details, including complex furniture and decorations. (2) Our dataset is rendered with GGX BRDF model~\cite{walter2007microfacet}, which has stronger material modeling capability than any BRDF models that existing indoor scene datasets use. (3) Our dataset provides not only pixel-wise surface environment maps, but also contains environment maps located anywhere in the 3D scene space. Fig.~\ref{fig:dataset} compares some example scenes in our dataset and OpenRooms, showing our dataset's higher scene quality.

\section{Network Design}

Our inverse rendering framework takes a single image of indoor scene as input and jointly predicts the spatially-varying material, geometry and lighting of the scene, and can further re-render the appearance of the input image. Fig.~\ref{fig:pipeline} overviews the architecture of our framework, which consists of three parts: material-geometry network (\S\ref{sec:mgnet}), lighting network (\S\ref{sec:light} and \S\ref{sec:background}), and a differentiable Monte-Carlo rendering layer (\S\ref{sec:render}). The material-geometry network is an end-to-end convolutional network which directly predict the reconstruction results. The lighting network $\mathbf{LightNet}$ is comprised of three sub-parts: A Resnet34 encoder to produce local feature map from the input image (like pixelNeRF~\cite{yu2021pixelnerf}), a screen-space ray tracer to trace the source of the light, and a final MLP decoder to predict the lighting radiance result. The rendering layer takes G-Buffers and lighting as input, and uses Monte Carlo raytracing to reproduce realistic rendering results.

\subsection{Material and Geometry Network}\label{sec:mgnet}
The input to our material and geometry prediction network $\mathbf{MGNet}$ is a single high dynamic range image, which can be directly obtained from our synthetic dataset. For real-world photos, we preprocess them with an inverse gamma correction. We use a single DenseNet121~\cite{huang2017densely} encoder to extract deep features of the material and shape parameters of the scene with different depth, as well as four separate decoders to obtain the final predicted albedo ($A$), material ($M$), normal ($N$), and depth ($D$), where $M$ consists of two parts: roughness $R$ and metallic $M_t$. While decoding neural features of different depths, upsampling and skip links are used to preserve multi-level details. Please refer to supplementary material for the detailed architecture of $\mathbf{MGNet}$. 

\subsection{Lighting Network}\label{sec:light}

We now describe our approach to predict any incident light intensity $L_i(\mathbf{p}, \mathbf{d})$ at point $\mathbf{p}$ with direction $\mathbf{d}$ from a single image. We fix our coordinate system as the \textit{view space} of the input image and specify position $\mathbf{p}$ and light direction $\mathbf{d}$ in this coordinate system.

Given an input image $\mathbf{I}$ of a scene, we first extract a feature map $\mathbf{F} = E(\mathbf{I})$, where $E$ is an encoder with ResNet34~\cite{he2016deep} architecture. For any location $\mathbf{x}$ in the scene, we can retrieve the corresponding image feature by projecting $\mathbf{x}$ onto the image coordinates $\pi(\mathbf{x})$ using camera intrinsics and extract the local feature vector $\mathbf{F}[\pi(\mathbf{x})]$. Instead of directly using the local feature at incident point $\mathbf{p}$, we trace the ray from $\mathbf{p}$ with direction $-\mathbf{d}$ to point $\mathbf{s}$ in the scene, which can be treated as a virtual point light of $L_i(\mathbf{p}, \mathbf{d})$. We extract the local feature vector $\mathbf{F}[\pi(\mathbf{s})]$. The local feature is then passed into the final MLP decoder $f$, along with view direction $\mathbf{d}$ and some local G-Buffers (diffuse albedo $K_d$, specular albedo $K_s$, normal $N$ and roughness $R$) at $\pi(\mathbf{s})$ $\mathbf{G}[\pi(\mathbf{s})]$, as
\begin{align}
    \mathbf{s} &= \mathrm{trace}(\mathbf{p}, -\mathbf{d}),\\
    L_i(\mathbf{p}, \mathbf{d}) &= f(\gamma(\mathbf{d}), \mathbf{F}[\pi(\mathbf{s})], \mathbf{G}[\pi(\mathbf{s})]),
\end{align}
where $\gamma(\cdot)$ is positional encoding function which is common used in NeRF~\cite{mildenhall2020nerf} to capture the high-frequency details within the data. The $\mathrm{trace}$ operation is implemented by \textbf{screen space ray tracing} (SSRT). We show our pipeline schematically in Fig.~\ref{fig:pipeline}.

Our \textbf{screen space ray tracer} works on the depth map of the scene. It takes depth map $\mathbf{D}$, starting point $\mathbf{p}$, and the tracing direction $\mathbf{d}$ as inputs. The screen space ray tracer performs ray marching through pixels from the start point. At each step, the current depth of the ray is updated and compared with the surface depth of the pixel. If the ray depth is larger, it indicates that the ray has passed through the pixel surface, i.e. an intersection has occurred. Otherwise, it continues ray marching to an adjacent pixel until hitting the edge of the image.

\subsection{Uncertainty-Aware Out-of-View Lighting Network }\label{sec:background}

A limitation of screen space ray tracing is that the traced ray does not necessarily intersect within the field of view of the image. Therefore, an additional network (named ``out-of-view lighting network'') is designed to handle lights from the out-of-view area of the scene. The design of our out-of-view lighting network is inspired by Neural Radiance Fields (NeRF)~\cite{mildenhall2020nerf}, which uses an MLP to represent the scene and uses volume rendering to predict the radiance of a ray. In the original version, the weights of the MLP are trained scene-specifically. Instead, we leverage hypernetwork~\cite{ha2016hypernetworks} to reconstruct out-of-view lighting by predicting the scene-specific weights of the NeRF MLP, and then query the radiance by the same volume rendering and alpha compositing techniques.

The out-of-view lighting network architecture is shown in Fig.~\ref{fig:pipeline} (right-hand side).
Given the input image $\mathbf{I}$, we first extract a global feature $\mathbf{F}_\mathrm{g} = G(\mathbf{I})$, where $G$ is an encoder with ResNet34 architecture (separate from the encoder in Section \ref{sec:light}). Then, $\mathbf{F}_\mathrm{g}$ is taken by hypernetwork $H$ and the MLP's weights $\Phi$ are returned. To query an incident light intensity $L_i(\mathbf{p}, \mathbf{d})$, we sample $N$ 3D points $\{\mathbf{x}_i = \mathbf{p} - t_i\mathbf{d}\}$ on ray $(\mathbf{p}, -\mathbf{d})$. With positional encoding $\gamma$ and NeRF MLP $f$, density $\sigma$ and RGB color $\mathbf{c}$ are returned. The complete process can be formulated as:
\begin{align}
    \Phi &= H(G(\mathbf{I})), \\
    \{\sigma_i, \mathbf{c}_i &= f(\gamma(\mathbf{x}_i); \Phi)\} _{i=1}^N .
\end{align}
Then light intensity $L$ can be composited by
\begin{equation}
    \hat{L} = \sum_{i=1}^N{T_i(1-\exp(-\sigma_i\delta_i))\mathbf{c}_i}, \text{where } T_i = \exp\left(-\sum_{j=1}^{i-1}{\sigma_j\delta_j}\right),
\end{equation}
where $\delta_i = t_{i+1} - t_i$ is the distance between adjacent samples. Due to performance consideration, our NeRF MLP is a small-scale network and does not take ray direction $\mathbf{d}$ as the MLP input like the original paper~\cite{mildenhall2020nerf} does.

The out-of-view lighting network is capable of predicting lighting anywhere in the scene. We now describe how we use it to refine the light predictions within field-of-view. Screen space ray tracing has a limitation that it may report some false positive of intersections. For real intersections, the difference between surface depth and ray depth is small, while the depth difference will increase when the intersection is a false positive. We model it as the ``uncertainty'' of SSRT, which is activated by hyperbolic tangent function: $u = \tanh(10\Delta d)$ where $\Delta d \in [0,\infty)$ is the depth difference. The refined light prediction is then formulated as 
\begin{equation}\label{eq:combine}
    \hat{L}_\mathrm{refined} = (1 - u) \times \hat{L}_\mathrm{SSRT} + u \times \hat{L}_\mathrm{out-of-view},
\end{equation}
where $\hat{L}_\mathrm{SSRT}$ is the light prediction by our SSRT-based lighting network and $\hat{L}_\mathrm{out-of-view}$ is the light prediction by our out-of-view lighting network. When uncertainty value $u$ is large, screen space ray tracing becomes untrusted and the final prediction is dominated by out-of-view lighting prediction. We ablate between using only out-of-view network predictions and using full model predictions combined with Eq.~\ref{eq:combine} in our supplementary material.
% \huo{How about changing the terminology 'background' to 'out-of-view' across the whole paper?}
\subsection{Rendering Layer}\label{sec:render}

\setlength\tabcolsep{0.5pt}
\begin{figure}[h]
    \centering
    \begin{tabular}{cccc}
        Ground Truth & \cite{li2020inverse} & Ours (no MC) & Ours \\
        \includegraphics[width=2.1cm]{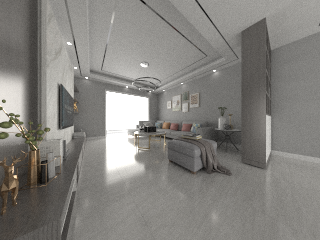} & \includegraphics[width=2.1cm]{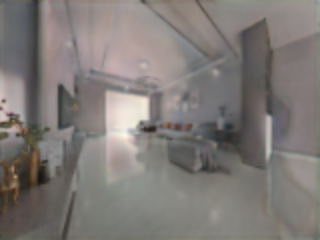} & \includegraphics[width=2.1cm]{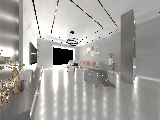} & \includegraphics[width=2.1cm]{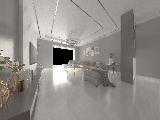} \\[-0.5ex]
        \includegraphics[width=2.1cm]{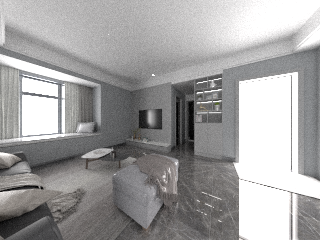} & \includegraphics[width=2.1cm]{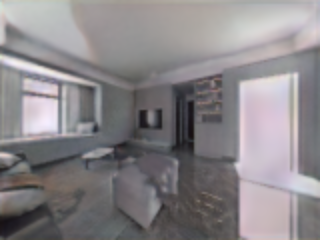} & \includegraphics[width=2.1cm]{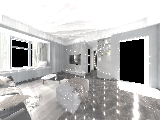} & \includegraphics[width=2.1cm]{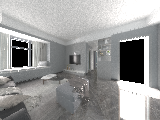} \\
    \end{tabular}
    \vspace*{-2.5ex}
    \caption{\textbf{Qualitative comparison on re-rendered image}. ``Ours (no MC)'' means that we re-render the image using our lighting prediction results but \citet{li2020inverse}'s rendering layer (instead of our MC rendering layer). Note that \citet{li2020inverse}'s render layer causes significant artifacts on glossy surfaces.}
    \label{fig:render}
    % \vspace*{-2.5ex}
\end{figure}

Unlike \citet{li2020inverse} which discretes the incident hemisphere to approximate the integration, we leverage differentiable Monte Carlo raytracing to produce photorealistic re-rendering results. Given sample count $N$, we use BRDF importance sampling to sample $N$ ray directions $\{d_i\} = \{\phi_i,\theta_i\}$ according to view direction, surface normal and material parameters (roughness and metallic) at pixel point $\mathbf{p}$.
We then perform screen-space raytracing according to $d_i$ to trace the source point and predict the radiance of the corresponding direction $\{\tilde{L_i}\}$ from $\mathbf{LightNet}$. The rendering layer computes the unbiased re-rendered image by
\begin{equation}\label{equ:render}
    \tilde{I} = \frac{1}{N}\sum_{i=1}^N{\frac{f_r(v,d_i;\tilde{A},\tilde{N},\tilde{R},\tilde{M_t})\tilde{L_i}\cos{\theta_i}}{p(v,d_i;\tilde{N},\tilde{R},\tilde{M_t})}},
\end{equation}
where $f_r(\omega_i,\omega_o)$ is the BRDF evaluation value and $p(\omega_i,\omega_o)$ is the probability distribution function (PDF) value of BRDF importance sampling, and $v$ is the view direction. Our importance sampling rendering layer can produce much more realistic re-rendered images compared to \cite{li2020inverse}, especially in specular reflections and highlights. As shown in Fig.~\ref{fig:render}, our rendering layer is capable of recovering specular reflections on the glossy floor, while the rendering layer used by \cite{li2020inverse} produces significant artifacts. The artifacts of \cite{li2020inverse}'s discretization rendering algorithm are caused by the deterministic discrete direction sampling at each pixel, which is likely to miss important directions in the specular BRDF term. The missing of important reflection directions results in interleaved patterns in the re-rendered result. In contrast, our importance sampling strategy can faithfully recover high-frequency reflections on glossy surfaces.

\section{Training}

We train our network models with the supervision of ground truth $\{I,A,N,D,R,M,L\}$ from our synthetic \textsc{InteriorVerse} dataset, where $A,N,D,R,M$ denote albedo, normal, depth, roughness, and metallic, respectively, and $L$ denotes spatially-varying lighting ground truth.

For geometry and material reconstruction, we use direct supervision to calculate the error between ground truth and network prediction. For lighting estimation, inspired by prior work~\cite{li2020inverse,wang2021learning}, to encourage photorealistic scene appearance reconstruction, we additionally use a differentiable in-network rendering layer to re-render the image according to the predicted material, geometry, and lighting, and try to recover the original input image through an image loss. Note that, unlike prior work, our render layer incorporates physically-based Monte Carlo sampling via screen space ray tracing, which explicitly regularizes the physical parameter space with GGX importance sampling. As we will demonstrate later, this makes our method significantly more robust to handle specular reflections in the interior scene. 

\subsection{Material-Geometry Network}

We train $\mathbf{MGNet}$ with the weighted combination of material losses (albedo loss $\mathcal{L}_{\mathrm{albedo}}$, roughness-metallic loss $\mathcal{L}_{\mathrm{material}}$) and geometry losses (normal loss $\mathcal{L}_{\mathrm{normal}}$ and depth loss $\mathcal{L}_{\mathrm{depth}}$):

\begin{equation}
    \mathcal{L}_\mathbf{MGNet} = \lambda_a\mathcal{L}_{\mathrm{albedo}} + \lambda_n\mathcal{L}_{\mathrm{normal}} + \lambda_m\mathcal{L}_{\mathrm{material}} + \lambda_d\mathcal{L}_{\mathrm{depth}}.
\end{equation}

We add perceptual loss~\cite{johnson2016perceptual} in the albedo, normal and material term, which helps to recognize the semantic boundaries in the image. The detailed definitions of separate losses and weights are presented in the supplemental material.

\subsection{Lighting Network}

We train $\mathbf{LightNet}$ with the weighted combination of direct light supervision loss $\mathcal{L}_\mathrm{light}$ and re-rendering loss $\mathcal{L}_{\mathrm{re-render}}$:

\begin{equation}
    \mathcal{L}_\mathbf{LightNet} = \mathcal{L}_\mathrm{light} + \lambda_r\mathcal{L}_{\mathrm{re-render}}
\end{equation}
where $\mathcal{L}_\mathrm{light}$ is the HDR supervision loss function proposed by \cite{mildenhall2021nerf}, while $\mathcal{L}_{\mathrm{re-render}}$ is an $L_2$ loss between the re-rendered image and the original image.
Please refer to supplementary material for the detailed definition of $\mathcal{L}_\mathrm{light}$ and $\mathcal{L}_{\mathrm{re-render}}$.

We find that re-rendering loss can significantly improve the lighting prediction, especially on specular surfaces. This benefit comes from enforcing the network to learn correct pixel brightness in $\hat{I}$, thus producing accurate lighting supervision in the scene and preventing blurry or spot artifacts in the re-rendered image. Ablation studies on the usage of re-rendering loss are presented in the supplementary material.

\subsection{Training Scheme}

We use a progressive training scheme to train our model in the order of data dependencies between different components of our framework. We first train material-geometry module to ensure correct predictions of albedo, normal, roughness, metallic and depth. This is because our lighting network depends on these properties (e.g., SSRT depends on depth, and MLP decoder depends on G-Buffers). Then we train lighting module with re-rendering loss.

\section{Experiments}\label{sec:experiments}

\subsection{Experiment Settings}

\paragraph{Training data.} We train our network on our new photorealistic indoor scene dataset,  introduced in Sec. \ref{sec:data}.  
When evaluating on real world data, we also fine-tune our model on IIW dataset~\cite{bell2014intrinsic} for albedo and NYUv2~\cite{silberman2012indoor} for depth and normal. Please refer to our supplementary material for more details on training and evaluation data.

\begin{figure}[ht]
    \centering
    % \vspace*{-2.5ex}
    \begin{tabular}{m{1em}<{\centering}m{2cm}<{\centering}m{2cm}<{\centering}m{2cm}<{\centering}m{2cm}<{\centering}}
          & \cite{li2020inverse} & Fine-tuned & Ours & Ground Truth \\[-0.2ex]
         \rotatebox{90}{Albedo} & \includegraphics[width=2cm]{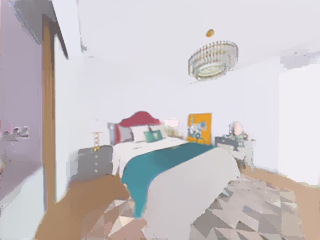} & \includegraphics[width=2cm]{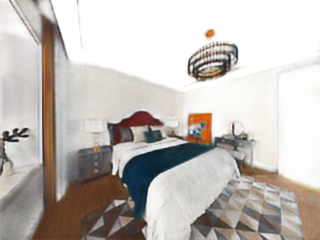}  & \includegraphics[width=2cm]{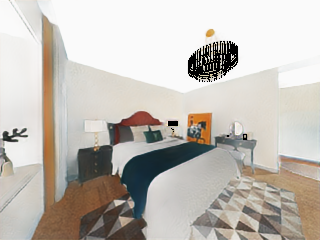} & \includegraphics[width=2cm]{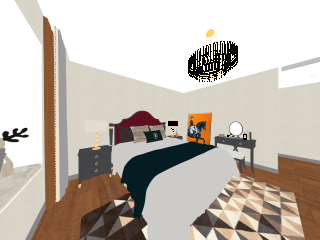}
         \\[-0.6ex]
         \rotatebox{90}{Roughness} & \includegraphics[width=2cm]{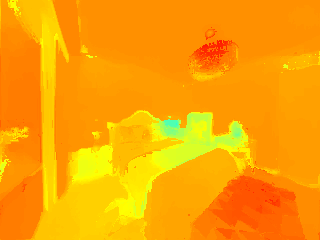} & \includegraphics[width=2cm]{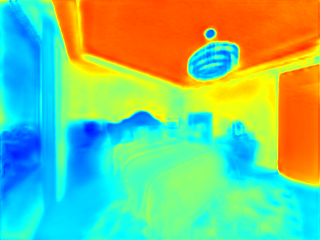}  & \includegraphics[width=2cm]{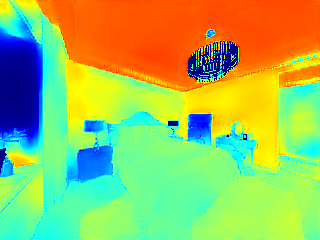} & \includegraphics[width=2cm]{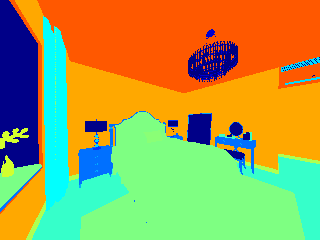} \\[-0.6ex]
         \rotatebox{90}{Normal} & \includegraphics[width=2cm]{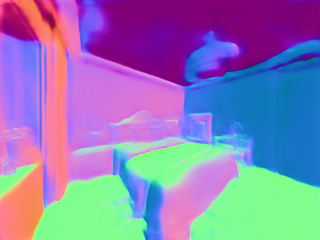} & \includegraphics[width=2cm]{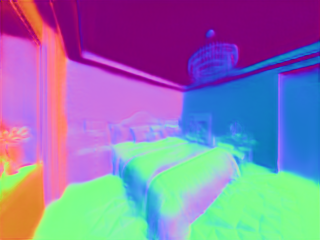}  & \includegraphics[width=2cm]{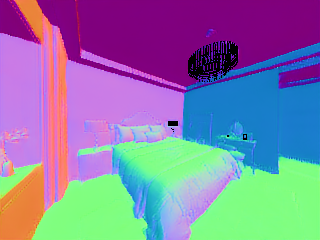} & \includegraphics[width=2cm]{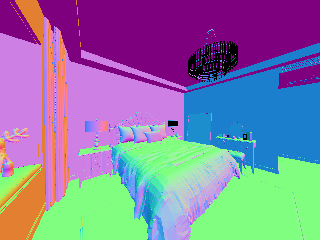} 
    \end{tabular}
    \vspace*{-2.5ex}
    \caption{\textbf{Qualitative results of geometry and BRDF estimation} on synthetic dataset between \citet{li2020inverse} and our method. In second row, we show the improved prediction of \citet{li2020inverse} by fine-tuning it on our \textsc{InteriorVerse} dataset. We omit metallic comparison since \citet{li2020inverse} does not support it. See supplementary for more results.}
    \label{fig:mgsyn}
    % \vspace*{-1.5ex}
\end{figure}

\paragraph{Baselines.} We compare our method with \citet{li2020inverse}, which is the state-of-the-art holistic inverse rendering frameworks for indoor scenes. To ensure a fair comparison, \emph{we fine-tune \cite{li2020inverse} on our new dataset}, which significantly improves its performance (Fig.~\ref{fig:mgsyn}). For lighting prediction, we compare with \cite{li2020inverse} as well as another state-of-the-art lighting estimation method Lighthouse~\cite{srinivasan2020lighthouse}, which requires a stereo image pair as input instead of a single image.

\setlength\tabcolsep{0.5pt}

\setlength\tabcolsep{1.2pt}
\begin{figure*}[ht]
    \centering
    \begin{tabular}{ccc|cc}
        Ground Truth & \cite{li2020inverse} & Ours & \cite{li2020inverse} & Ours \\[-0.1ex]
        
        \includegraphics[width=0.191\textwidth]{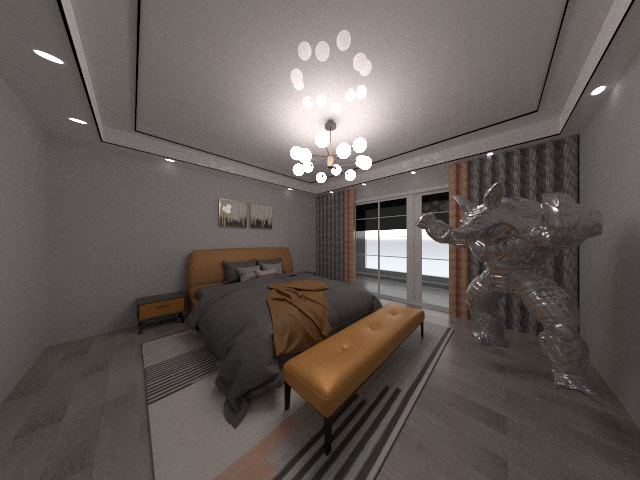}&
        \includegraphics[width=0.191\textwidth]{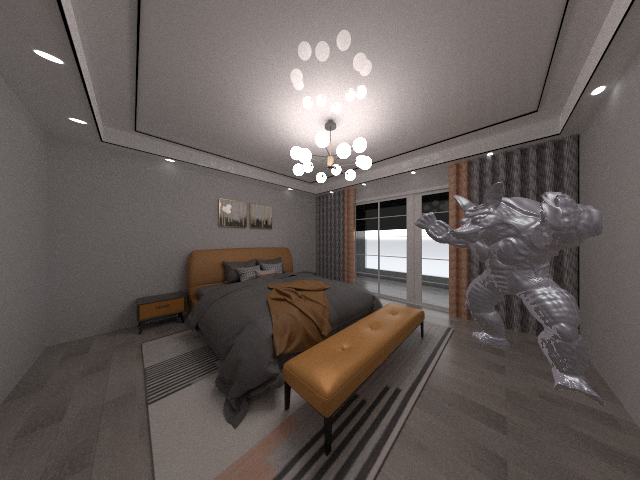}&
        \includegraphics[width=0.191\textwidth]{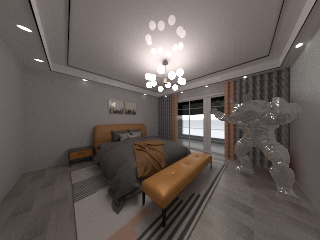}&
        \stackinset{r}{0pt}{b}{0pt}{\includegraphics[width=0.05\textwidth]{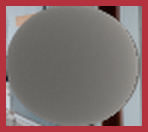}}{\includegraphics[width=0.191\textwidth]{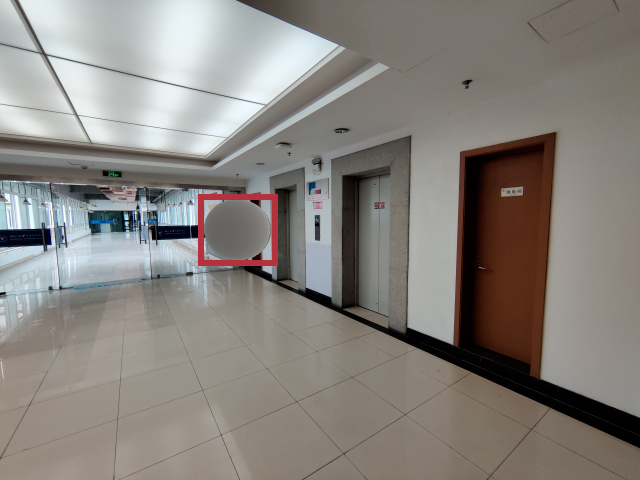}}&
        \stackinset{r}{0pt}{b}{0pt}{\includegraphics[width=0.05\textwidth]{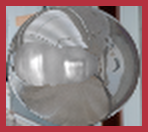}}{\includegraphics[width=0.191\textwidth]{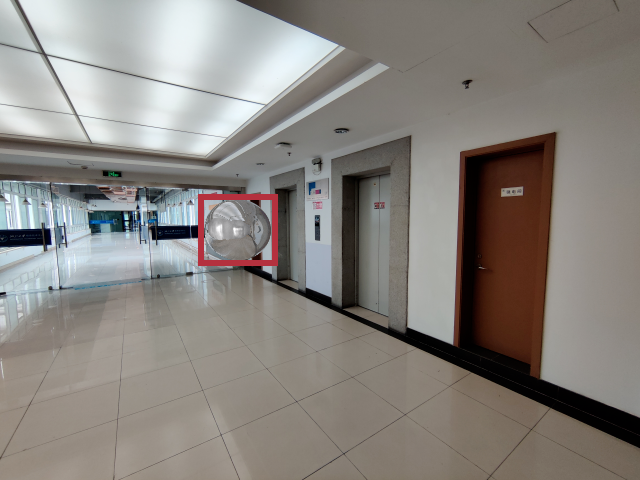}} \\[-0.35ex]
        
        \includegraphics[width=0.191\textwidth]{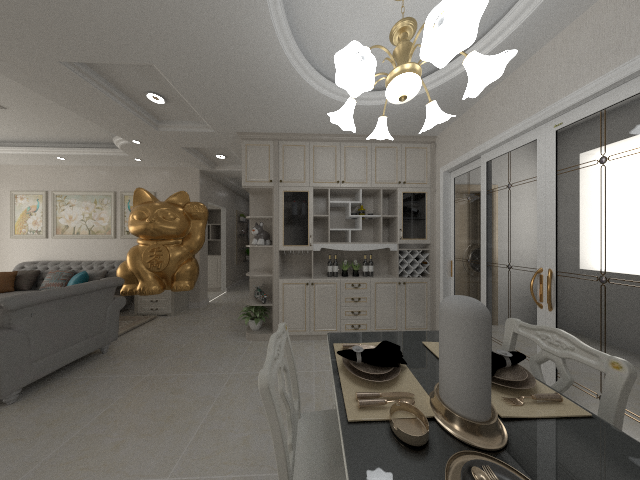}&
        \includegraphics[width=0.191\textwidth]{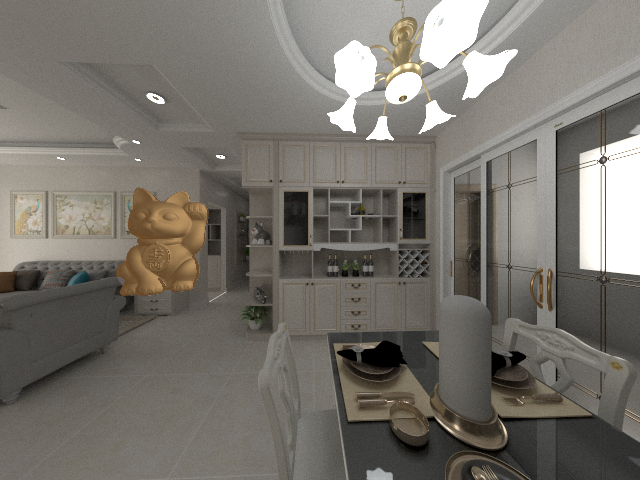}&
        \includegraphics[width=0.191\textwidth]{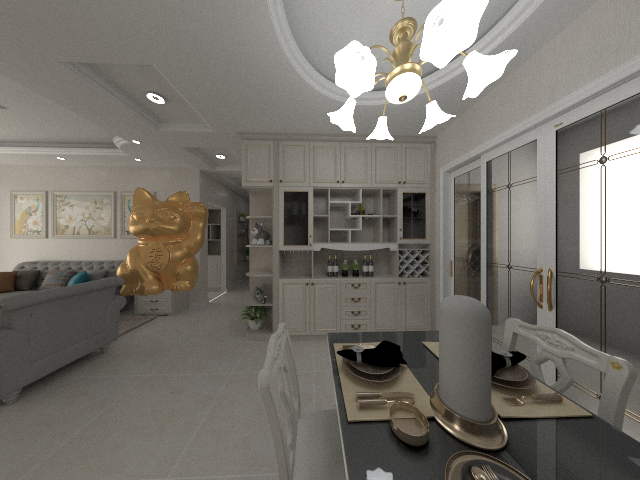}&
        \stackinset{r}{0pt}{b}{0pt}{\includegraphics[width=0.05\textwidth]{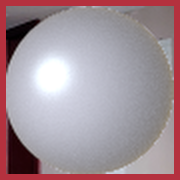}}{\includegraphics[width=0.191\textwidth]{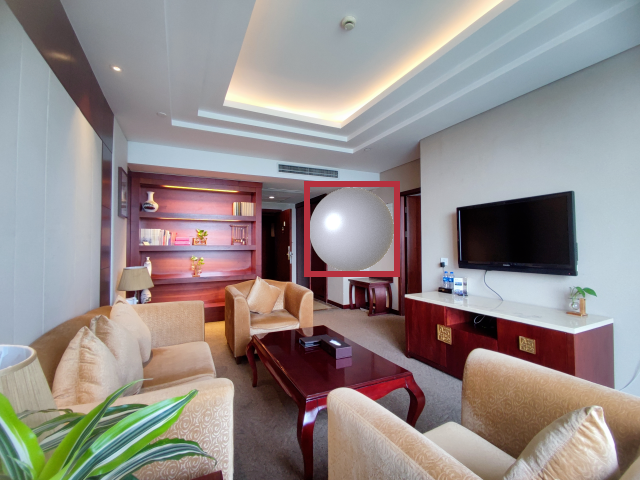}}&
        \stackinset{r}{0pt}{b}{0pt}{\includegraphics[width=0.05\textwidth]{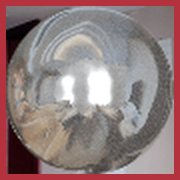}}{\includegraphics[width=0.191\textwidth]{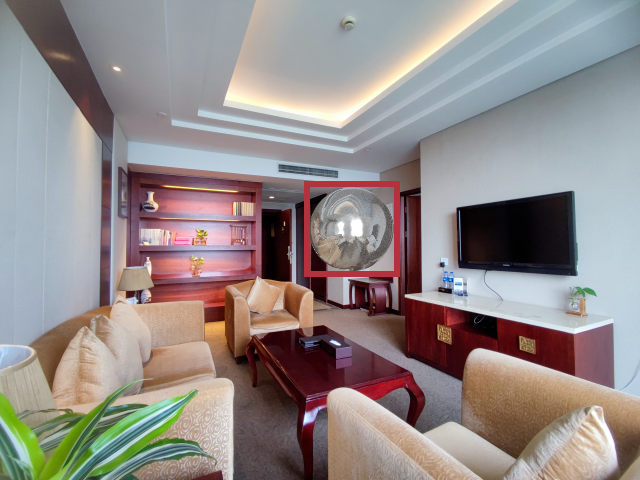}} \\[-0.35ex]
        
        &\textsc{Synthetic data}&&
        \multicolumn{2}{c}{\textsc{Real data} } 
    \end{tabular}
    \vspace*{-1.5ex}
    \caption{\textbf{Qualitative comparison of object insertion} results on synthetic dataset and real-world images. The ground truth object insertion results of synthetic scenes are provided. Li et al.'s results use the same highly-specular GGX BRDF as our results. However, because of their low-frequency lighting prediction, the inserted objects contain no sharp reflections and therefore resemble Lambertian appearance.}
    \label{fig:insert_real}
\end{figure*}

\subsection{Evaluation of Material and Geometry}

We evaluate material (albedo, roughness, and metallic) and geometry (normal and depth) prediction on \textsc{InteriorVerse} synthetic indoor dataset, as well as real-world dataset (NYUv2 dataset~\cite{silberman2012indoor} for geometry and IIW dataset~\cite{bell2014intrinsic} for albedo).

\paragraph{Evaluation on synthetic dataset.} We compare our method with the baseline methods on our \textsc{InteriorVerse} dataset. As shown in Fig.~\ref{fig:mgsyn}, our method outperforms \cite{li2020inverse},
For albedo prediction, while \cite{li2020inverse} tends to over-smooth the result, our method faithfully preserves the texture details (e.g., the wooden textures of the floor). For normal prediction, our method is capable of preserving sharp edges between walls and floors. This attributes to the usage of perceptual loss, which helps the model recognize semantic borders in the image. Please refer to the supplementary material for an ablation study on the usage of perceptual loss.

\paragraph{Evaluation on real-world datasets.} We evaluate albedo prediction on IIW dataset~\cite{bell2014intrinsic} with sparse pairwise human albedo annotations. We use the official metric suggested by~\cite{bell2014intrinsic}, Weighted Human Disagreement Rate (WHDR), which measures the error when albedo predictions disagree with human annotations.
We also evaluate geometry prediction on NYUv2 dataset~\cite{silberman2012indoor}. As shown in Table \ref{tab:nyuv2_and_iiw}, we observe a lower error compared to prior works~\cite{li2020inverse,wang2021learning}, indicating the advantage of our photo-realistic training datasets and our network design.
Qualitative results of geometric and material predictions on real-world data are presented in the supplementary material.

\begin{table}[h]
    \centering
    % \vspace*{-2.5ex}
    \caption{\textbf{Evaluation of normals and depth} on NYUv2 dataset (2nd and 3rd columns), and \textbf{albedo} on IIW dataset (last column).}
    \vspace*{-2.5ex}
    \begin{tabular}{|c|c|c|c|}
    \hline
       Method & Normal Angular Error & Depth si-MSE & WHDR \\\hline
        \cite{li2020inverse} & $24.12^{\circ}$ & 0.160 & 15.9 \\\hline
        \cite{wang2021learning} & $22.95^{\circ}$ & 0.181 & 18.2\\\hline
        Ours & $\mathbf{21.86^{\circ}}$ & \textbf{0.155} & \textbf{15.5} \\\hline
    \end{tabular}
    \label{tab:nyuv2_and_iiw}
\end{table}

\subsection{Evaluation of Lighting}

\paragraph{Evaluation on virtual object insertion.} We evaluate our lighting estimation method on a crucial augmented reality application: virtual object insertion. With the help of screen space ray tracing and the Monte Carlo rendering layer, we can achieve promising results in specular reflection effects. Fig.~\ref{fig:insert_real} shows results of our method compared to baselines, consisting of both synthetic data and real world images. In order to emphasize the ability to recover high-frequency lighting details, the materials of the inserted objects are \emph{highly specular}. For synthetic data, we insert complex objects and ground truths are provided. \citet{li2020inverse}'s lighting estimation is 2D spatially-varying, which cannot handle 3D points far from 2D surfaces. Moreover, their Spherical Gaussian lighting representation is incapable of capturing high-frequency angular details. Therefore, the appearance of inserted highly specular objects does not contain sharp reflections. In contrast, our method produces photorealistic shading and specular highlights on the inserted object. For real world data, we choose to insert highly specular spheres. The reflection on the sphere is supposed to be consistent with the surrounding environments. \citet{li2020inverse} also fails in this task, due to its low-frequency lighting predictions, while our method manages to faithfully recover angular details of the surrounding environment on the inserted sphere.

\setlength\tabcolsep{0.5pt}
\begin{figure}[h]
    \vspace*{-1.5ex}  
    \centering
    \begin{tabular}{ccc}
        Origin & Lighthouse~\shortcite{srinivasan2020lighthouse} & Ours \\
        \includegraphics[width=2.8cm]{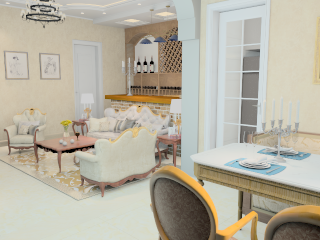} & 
        \stackinset{r}{0pt}{b}{0pt}{\includegraphics[width=0.8cm]{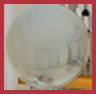}}{\includegraphics[width=2.8cm]{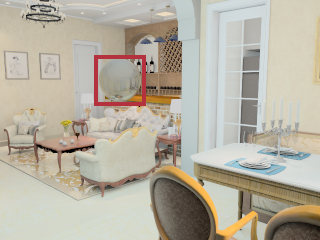}} &
        \stackinset{r}{0pt}{b}{0pt}{\includegraphics[width=0.8cm]{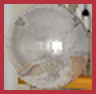}}{\includegraphics[width=2.8cm]{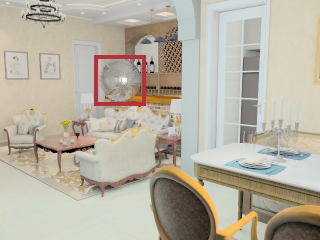}}
    \end{tabular}
    \vspace*{-2.5ex}    
    \caption{\textbf{Qualitative comparison of object insertion} results between Lighthouse~\shortcite{srinivasan2020lighthouse} and our method on \textbf{Lighthouse's test set}.}
    \vspace*{-1.5ex}
    \label{fig:lighthouse}
\end{figure}

We also compare our method with another state-of-the-art lighting estimation method Lighthouse~\cite{srinivasan2020lighthouse}, which requires a stereo pair of images as input. To show our method's cross-domain ability, we evaluate on \textit{Lighthouse's test set} from InteriorNet~\cite{li2018interiornet} \textit{without fine-tuning our network}. As shown in Figure~\ref{fig:lighthouse}, our method outperforms Lighthouse, even with a lower number of input images and a potential domain gap. We can observe that Lighthouse's lighting prediction has significantly less variation in lighting intensity. This may be because Lighthouse is trained from LDR panoramas, and cannot handle HDR lighting.

We also explore more applications of our lighting estimation method, including re-rendering and scene material edit. Please refer to our supplementary material for these additional results.

\section{Conclusion and Limitations}
We present a learning-based method for inverse rendering of complex indoor scenes. Our approach handles spatially-varying illumination and faithfully recovers specular reflections thanks to the differentiable Monte Carlo rendering layer, enabling photorealistic editing such as complex object insertion and material change. Lastly, we introduce a large-scale indoor dataset, \textsc{InteriorVerse}, which contains much richer details than existing alternatives. 

There are some limitations of our method. Our out-of-view lighting network is not capable of predicting high-frequency details due to its limited network capacity. Monte Carlo sampling would also lead to noisy re-render results, and raising the required sample budget can be computationally expensive. Further, emission of light sources is not supported currently, which we leave as future work.

% \begin{acks}
% This work was supported in part by Key R\&D Program of Zhejiang Province (No. 2022C01025), NSFC (No. 61872319), the Fundamental Research Funds for the Central Universities, Zhejiang Lab (121005-PI2101), and Information Technology Center and State Key Lab of CAD\&CG, Zhejiang University.
% \end{acks}

\newpage

%%
%% The next two lines define the bibliography style to be used, and
%% the bibliography file.
\bibliographystyle{ACM-Reference-Format}
\bibliography{ref}

% %%
% %% If your work has an appendix, this is the place to put it.
% \appendix

\end{document}